\documentclass{article}
\usepackage{amssymb}
\usepackage{graphicx}
\usepackage{adjustbox}
\usepackage{booktabs} 
\usepackage{color}
\usepackage{amsmath}
\usepackage{caption}
\usepackage[table]{xcolor} 
\usepackage{multirow}
\usepackage{wrapfig}
\usepackage{graphicx}
\usepackage{subcaption}

\definecolor{ceiling}{RGB}{214, 8, 40}   %
\definecolor{floor}{RGB}{43, 160, 4}     %
\definecolor{wall}{RGB}{158, 216, 229}  %
\definecolor{window}{RGB}{114, 158, 206}  %
\definecolor{chair}{RGB}{204, 204, 91}   %
\definecolor{bed}{RGB}{255, 186, 119}  %
\definecolor{sofa}{RGB}{147, 102, 188}  %
\definecolor{table}{RGB}{30, 119, 181}   %
\definecolor{tvs}{RGB}{160, 188, 33}   %
\definecolor{furniture}{RGB}{255, 127, 12}  %
\definecolor{objects}{RGB}{196, 175, 214} %

\usepackage[preprint]{corl_2025} 

\title{RoboOcc: Enhancing the Geometric and Semantic Scene Understanding for Robots}

\author{
    Zhang Zhang$^{1,2,*}$,
    Qiang Zhang$^{1,3,*}$,
    Wei Cui$^{1,*}$,\\
    Shuai Shi$^{1}$,
    Yijie Guo$^{1}$,
    Gang Han$^{1}$,
    Wen Zhao$^{1}$,
    Hengle Ren$^{1}$,
    Renjing Xu$^{3}$,
    Jian Tang$^{1,\dagger}$\\
    $^{1}$ Beijing Innovation Center of Humanoid Robotics \\
    $^{2}$ Beijing Institute of Technology \\
    $^{3}$ Hong Kong University of Science and Technology (Guangzhou) \\
    \footnotesize{$^*$ Contributed equally.}
    \footnotesize{$^\dagger$ Corresponding author.}
}

\begin{document}
    \renewcommand\twocolumn[1][]{#1}%
    \maketitle
    \vspace{-5mm}
    \begin{center}
        \centering
        \includegraphics[width=\linewidth]{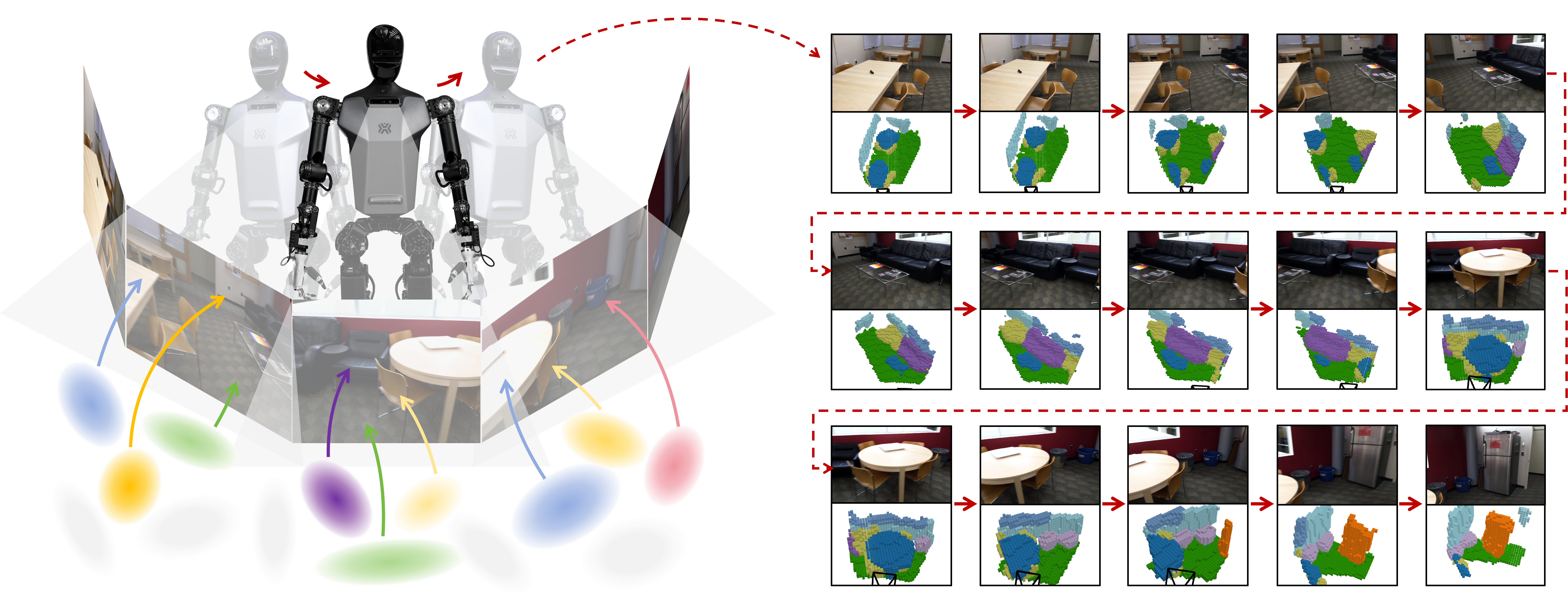}
        \vspace{-5mm}
        \captionof{figure}{Considering that current Gaussian representations lack effective utilization of geometry and opacity properties, we propose an enhanced geometric and semantic scene understanding 3D occupancy prediction method for robots. Based on this, the robot makes the local occupancy prediction in an indoor scene with accepted monocular RGB and completes the global occupancy prediction through exploration over time.
    }
    \label{fig_1}
    \end{center}


\begin{abstract}
    3D occupancy prediction enables the robots to obtain spatial fine-grained geometry and semantics of the surrounding scene, and has become an essential task for embodied perception. Existing methods based on 3D Gaussians instead of dense voxels do not effectively exploit the geometry and opacity properties of Gaussians, which limits the network's estimation of complex environments and also limits the description of the scene by 3D Gaussians. In this paper, we propose a 3D occupancy prediction method which enhances the geometric and semantic scene understanding for robots, dubbed RoboOcc. It utilizes the Opacity-guided Self-Encoder (OSE) to alleviate the semantic ambiguity of overlapping Gaussians and the Geometry-aware Cross-Encoder (GCE) to accomplish the fine-grained geometric modeling of the surrounding scene. We conduct extensive experiments on Occ-ScanNet and EmbodiedOcc-ScanNet datasets, and our RoboOcc achieves state-of the-art performance in both local and global camera settings. Further, in ablation studies of Gaussian parameters, the proposed RoboOcc outperforms the state-of-the-art methods by a large margin of (8.47, 6.27) in IoU and mIoU metric, respectively. The codes will be released soon.
\end{abstract}



\section{Introduction}
    \vspace{-2mm}
    The rise of embodied intelligence \cite{wu2024embodiedocc, zhang2025humanoidpano, liu2024volumetric, wang2024embodiedscan, xu2024embodiedsam} and computer vision \cite{qi2019deep, liu2024aligning, qi2020imvotenet, couprie2013indoor, rozenberszki2022language, wald2019rio, jia2024sceneverse} draws vast attention to 3D scene understanding, which enables robots to explore environments, make decisions, and carry out a range of downstream tasks. In 3D scene understanding, the 3D occupancy prediction method captures arbitrarily shaped obstacles by predicting the occupancy state of each voxel in the surrounding 3D space, which demonstrates its robustness, uniformity and scalability when facing complex environments. Existing methods \cite{wu2024embodiedocc, huang2024gaussianformer} employ 3D Gaussians rather than dense voxels as the flexible scene representations, making significant improvements on indoor 3D occupancy prediction. However, previous works \cite{huang2024gaussianformer, wu2024embodiedocc} do not effectively exploit the geometry and opacity properties of Gaussians. The lack of effective participation of geometry properties prevents the network from effectively obtaining spatial information, making it difficult to accomplish fine-grained modeling of indoor scenes, and the neglect of opacity properties largely triggers semantic ambiguities in prediction, resulting in degraded performance. 

    \begin{wrapfigure}{l}{0.5\textwidth}
    \centering
    \includegraphics[width=\linewidth]{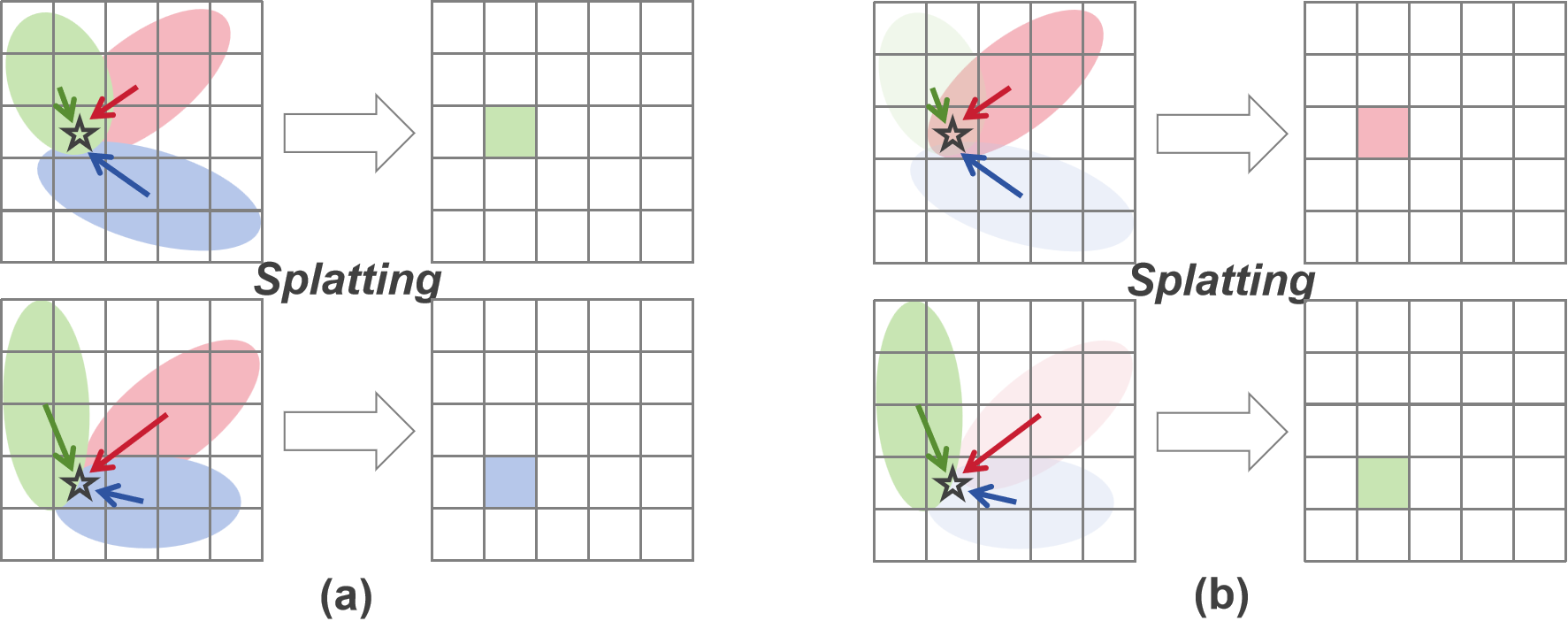}
    \caption{\textbf{The pipeline of Gaussian-to-Voxel splatting.} The green, pink, blue ellipsoid and grid represents different 3D Gaussian and voxel in 3D space, respectively. The arrow points from Gaussian center to voxel. The comparison will be conducted under controlled variable conditions.}
    \label{fig_2}
    \end{wrapfigure}
    
    The geometry properties of 3D Gaussians are primarily involved in the interaction of Gaussian queries with image features and act on the Gaussian-to-Voxel splatting which generates the final occupancy. Specifically, for each 3D Gaussian, it generates a series of sampling points based on its ellipsoid geometry, which are projected onto the image for deformable attention. The 3D Gaussians occupy voxels based on their geometry and ends up contributing semantics to each voxel occupied by it in 3D space. As shown in Figure \ref{fig_2} (a), the Gaussian whose center is closer to the voxel will contribute more semantics. In sparse and heavily empty-occupied outdoor scenes, coarse-grained feature sampling and modeling affects the quality of 3D occupancy predictions to a relatively small extent, which is not the case in indoor scenes. Indoor scenes have finer object classification and more accurate shape descriptions than outdoor scenes, and its associated downstream tasks also require more fine-grained modeling. However, geometry properties are not currently effectively involved in updating Gaussian queries. The sampling points generated based on their geometry could only sample image features that correspond to their spatial locations, but do not enable the interaction process to truly perceive Gaussian shapes. The 3D Gaussians that are not effectively refined for elliptical geometry produce invalid occupancy and coarse-grained modeling in the final splatting. 
    
    On the other hand, the opacity properties of the 3D Gaussian are weighted against the Gaussian around the voxel in Gaussian-to-Voxel splatting which generates the final semantic occupancy prediction. Specifically, for each voxel in 3D space, it searches for all 3D Gaussians that occupy it and weights the semantic properties of the 3D Gaussians, which is used to generate the final semantic category of the voxel. As shown in Figure \ref{fig_2} (b), the 3D Gaussians with higher opacity will contribute more semantics. However, opacity is not currently effectively involved in the process of updating Gaussian queries, which induces a semantic ambiguity for overlapping Gaussians in the Gaussian-to-Voxel splatting. Furthermore, in ablation studies of Gaussian parameters, we found that the network produces severe performance degradation due to the Gaussian overlaps. Specifically, we conduct the experiments by reducing the number of Gaussians and increasing the max scale of Gaussians, which causes a larger degree of Gaussian overlaps. The semantic ambiguity of the generated voxels is further exacerbated when a larger degree of Gaussian overlap meets 3D Gaussians with coarse-grained geometry and opacity confusion.
    
    In this work, we propose a 3D occupancy prediction approach which enhances the geometric and semantic scene understanding for robots, dubbed RoboOcc, as shown in Figure \ref{fig_1}. It utilizes the Opacity-guided Self-Encoder (OSE) to alleviate the semantic ambiguity of overlapping Gaussians and the Geometry-aware Cross-Encoder (GCE) to accomplish the fine-grained geometric modeling of the surrounding scene. The proposed RoboOcc effectively obtains the 3D semantic Gaussians from image inputs by three steps. First, it randomly initializes a set of 3D semantic Gaussians to sparsely describe a 3D scene. Each Gaussian represents a flexible region of interest and consists of the geometry, opacity and its semantic category. The 3D Gaussians expand the receptive field of 3D space via sparse convolution and focuses on the features of the non-empty Gaussian via opacity-guided gated convolution. Second, we generate a series of sampling points based on the ellipsoid geometry and project them to the multiscale image features. The cross-attention mechanism is performed to aggregate information. We enhance the network's ability to capture key sampling point features via semantic mixing and accomplish fine-grained modeling via geometry mixing. Third, we decode and predict new Gaussian properties from the updated Gaussian queries, consisting of the geometry, opacity and its semantic category.
    
    We update and optimize the 3D Gaussians by iterating the above self-encoder, cross-encoder and decoder. Finally, we aggregate the neighboring Gaussians to generate the semantic occupancy for a certain 3D voxel position via Gaussian-to-Voxel splatting. We conduct extensive experiments on the Occ-ScanNet and EmbodiedOcc-ScanNet datasets for 3D semantic occupancy prediction from local and global cameras, respectively. The proposed RoboOcc outperforms the state-of-the-art methods in both local and global camera settings. Further, in ablation studies of Gaussian parameters, the proposed RoboOcc outperforms the state-of-the-art methods by a large margin of (8.47, 6.27) in IoU and mIoU metric, respectively.

\vspace{-2mm}
\section{Related Work}
\label{sec:related_work}
\subsection{3D Occupancy Prediction}
    In recent years, 3D occupancy prediction \cite{huang2023tri, cao2022monoscene, wei2023surroundocc} has received increasing attention for its comprehensive and flexible description of indoor and outdoor scenes \cite{silberman2012indoor, dai2017scannet, dai2018scancomplete, tian2023occ3d, wang2023openoccupancy}, and has been extended to a range of downstream tasks \cite{hu2023planning, zheng2024occworld, wang2024occsora}, advancing the development of intelligent agents. MonoScene \cite{cao2022monoscene} proposed the first framework for direct prediction of 3D semantic occupancy from monocular images and provided inspiration for follow-up works. Subsequently, it becomes crucial to represent the surrounding 3D scene effectively and efficiently. Many recent works \cite{chen20203d, jiang2024symphonize, li2023fb, zhang2023occformer} have demonstrated the strengths by directly discretizing the 3D space into regular voxels and using dense representations of voxels to structure the 3D scene. Nonetheless, methods based on dense voxels ignored the spatial sparsity of the environment, leading to inherent redundancy.

\subsection{Object-centric Scene Representation}
    High resolution voxels make it challenging to compute in real time for the voxel-based 3D occupancy prediction methods. Subsequent works \cite{tang2024sparseocc, li2023voxformer} turned to sparse object-centric representation as a solution to above problem. However, non-empty areas may be incorrectly categorized as unoccupied and excluded completely throughout the subsequent process. GaussianFormer \cite{huang2024gaussianformer} proposed the object-centric 3D scene representation for 3D semantic occupancy prediction where each unit describes a region of interest instead of fixed grids or sparse voxels. It utilizes a series of spatial 3D Gaussians for a flexible representation that preserves the sparsity of occupancy space and the diversity of object scales, making important advances. Subsequent work \cite{wu2024embodiedocc} extended it to the continuous perception of indoor scenes, which is progressively accomplished through an incremental perception approach. However, previous methods ignored the necessity of fine-grained modeling of indoor scenes, and at the same time, their ignorance of the opacity properties of the 3D Gaussians will largely trigger semantic ambiguities in the prediction results, resulting in degraded performance.

\vspace{-2mm}
\section{Method}
\label{sec:method}
\subsection{Problem Formulation}
    In this work, we aim to obtain the local 3D occupancy prediction from indoor monocular RGB image within the current frustum and the global 3D occupancy prediction from historical frame information. The local branch is as shown in below:
    \begin{equation}
    O_{local} = F_{local}(I, E, K) 
    \label{eq1}
    \end{equation}
    
    Formally, we are given the RGB image $I\in \mathbb{R}^{H \times W \times 3}$ from the indoor monocular camera, where the $\{H,W\}$ denotes the image resolution. The extrinsic matrix $E\in \mathbb{R}^{3 \times 4}$ and intrinsic matrix $K\in \mathbb{R}^{3 \times 3}$ can be obtained via camera calibration. The $F_{local}$ is the proposed local prediction model, and the $O_{local} \in \mathbb{R}^{X_{local} \times Y_{local} \times Z_{local} \times C_{class}}$ is the local 3D occupancy prediction, where the $\{X_{local},Y_{local},Z_{local}\}$ and $C_{class}$ denote the target volume resolution of the local front view and the set of semantic classes.

    When we obtain the 3D occupancy prediction for the current view and move to the next viewpoint, the global branch is as shown in below:
    \begin{equation}
    O_{t} = F_{global}(I_{t}, E_{t}, K_{t}, O_{t-1})
    \label{eq2}
    \end{equation}

    Given the RGB image $I$, extrinsic matrix $E$, intrinsic matrix $K$ at the current timestamp $t$ and history 3D occupancy prediction $O_{t-1}$ preserved at the timestamp $t-1$ of the global scene, we obtain the 3D occupancy prediction $O_{t} \in \mathbb{R}^{X_{global} \times Y_{global} \times Z_{global} \times C_{class}}$ preserved at the current timestamp $t$ of the global scene via the global prediction model $F_{global}$, where the $\{X_{global},Y_{global},Z_{global}\}$ denotes the target volume resolution of the global scene.

\subsection{RoboOcc}
    \textbf{Overall Architecture.} As shown in Figure \ref{fig_3}, the overall RoboOcc framework consists of Image Encoder, Gaussian Encoder and Gaussian-to-Voxel Splatting. For Image Encoder, We use the EfficientNet to extract multi-scale semantic features from monocular image. We then randomly initialized a set of Gaussian queries and anchors. We use the 3D Gaussians to represent indoor scene and update the Gaussian-based representation based on semantic and structural features extracted from indoor monocular image with Gaussian Encoder. The Gaussian-to-Voxel splatting is finally employed to generate dense 3D occupancy prediction via local aggregation of Gaussians.

    \textbf{Gaussian Initialization.} We use the 3D Gaussians to represent indoor scene and update the Gaussian-based representation based on semantic and structural features extracted from indoor monocular image. Specifically, we use a set of 3D Gaussian anchors $A \in \mathbb{R}^{N \times D}$ and 3D Gaussian queries $Q \in \mathbb{R}^{N \times C}$ for each scene, where $N$ and $C$ denote the number of 3D Gaussian anchors and channel dimension of 3D Gaussian queries. Each 3D Gaussian anchor is represented by a D-dimensional vector in the form of $\{m \in \mathbb{R}^{3}, s\in \mathbb{R}^{3}, r \in \mathbb{R}^{4}, o \in \mathbb{R}^{1}, c \in \mathbb{R}^{C}\}$, where $m$, $s$, $r$, $o$ and $c$ denote the mean, scale, rotation, opacity vectors and semantic categories, respectively. Each 3D Gaussian query is projected into the depth map generated by the frozen DepthAnything model based on its mean coordinate property to obtain the pixel-aligned depth feature. Then, we use the Gaussian Encoder to facilitate the Gaussian self-interaction, Gaussian-to-image cross-interaction and updates between Gaussians. Next we will introduce the self-interaction, cross-interaction and update step-by-step.

    \textbf{Opacity-guided Self-Encoder.} Self-encoders based on sparse convolution are receptive field constrained and unable to effectively discriminate between foreground and background Gaussians due to the lack of opacity involved. Instead, we use opacity-guided gated sparse convolution to make the network more attentive to non-empty Gaussians and the multi-scale module to expand the spatial receptive field, as shown in Figure \ref{fig_3}. The opacity-guided gated sparse convolution provides the guidance for the foreground-background correction of the self-encoder.

    We first compose the sparse tensor representation with the mean property of Gaussian anchors and the features of Gaussian queries. Specifically, we treat each Gaussian mean as a point, generate the point cloud and voxelize it, and the features of Gaussian queries are the features of the point cloud.

    The proposed opacity-guided gated sparse convolution $OGSPConv$ is as shown in below:
    \begin{equation}
    OGSPConv(Q, o) = SPConv_{3\times3}(Q) \odot (Sigmoid(o)+Sigmoid(SPConv_{1\times1}(Q)))
    \label{eq3}
    \end{equation}
    Given the Gaussian queries $Q$ and Gaussian properties $o$, we utilize the sigmoid function to obtain a weighted foreground-background score, where the $Sigmoid$ denotes the sigmoid function, the $SPConv_{1\times1}$ and the $SPConv_{3\times3}$ denote the submanifold sparse convolution function with kernel size $1\times1$ and $3\times3$. We then multiply the score with the Gaussian queries element-by-element to get the opacity-guided Gaussian queries output.



    \begin{figure*}[!t]
    \centering
        \includegraphics[width=1\textwidth]{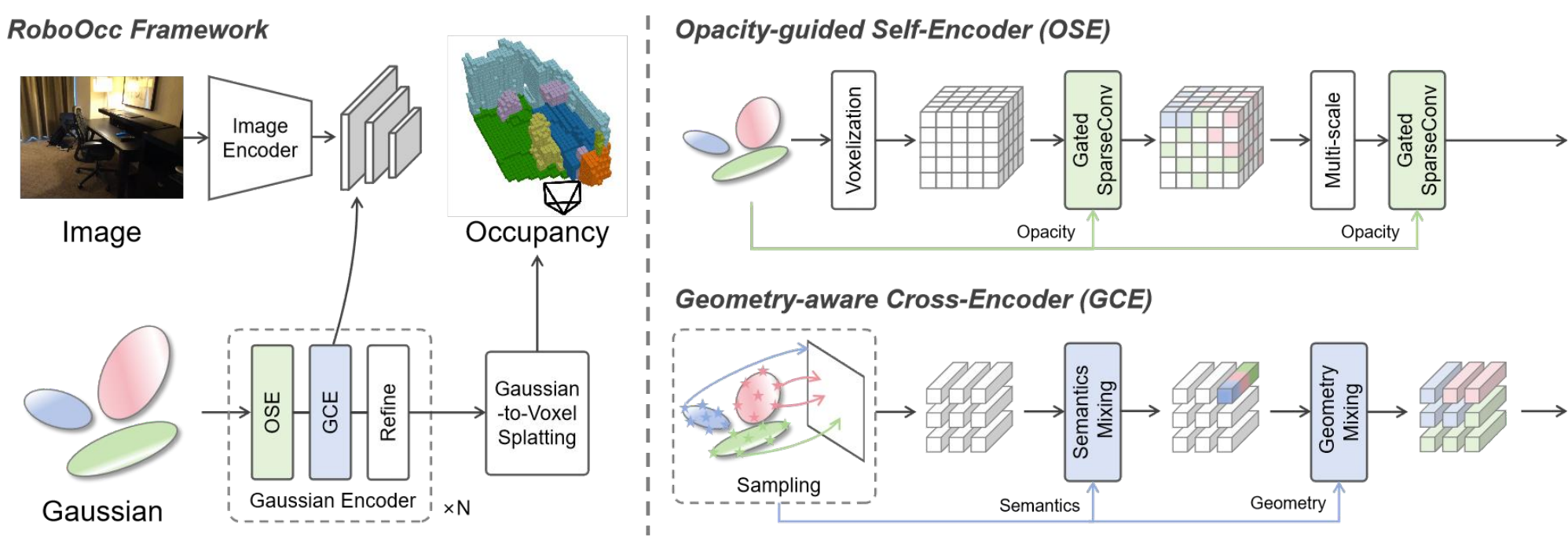}
        \caption{\textbf{The overall framework of the proposed RoboOcc.} It consists of Image Encoder, Gaussian Encoder and Gaussian-to-Voxel Splatting. For Image Encoder, We use the EfficientNet to extract multi-scale semantic features from monocular image. We then randomly initialized a set of Gaussian queries and anchors. We use the 3D Gaussians to represent indoor scene and update the Gaussian-based representation based on semantic and structural features extracted from indoor monocular image with Gaussian Encoder. The Gaussian-to-Voxel splatting is finally employed to generate dense 3D occupancy prediction via local aggregation of Gaussians.}
        \vspace{-2mm}
        \label{fig_3}
    \end{figure*}
    
    \textbf{Geometry-aware Cross-Encoder.} The cross-encoder is designed to extract semantic and structural features from indoor monocular image features. We initialize a series of offsets $\Delta m_0 \in \mathbb{R}^{N \times R}$, where the $N$ and $R$ denote the number of 3D Gaussians and offsets. The initialized offsets are then used to obtain Gaussian geometric offsets $\Delta m \in \mathbb{R}^{N \times R}$ by multiplying them with the scale properties $s$ and rotation properties $r$ in the Gaussian geometry properties, as shown in below:
    \begin{equation}
    \Delta m = MatMul(r, \Delta m_0 s)
    \label{eq4}
    \end{equation}
    Where the $MatMul$ denotes the matrix multiplication. A series of ellipsoidal reference points $P \in \mathbb{R}^{N \times R}$ are generated by adding the Gaussian geometric offsets to the mean properties $m$ of each Gaussian. Then we project the 3D reference points onto image feature maps to get the sampled queries $Q_p$ with extrinsic matrix $E$, intrinsic matrix $K$, as shown in below:
    \begin{equation}
    Q_p = Sampling(Q, \pi(P, E, K), F)
    \label{eq5}
    \end{equation}
    Where the $Sampling$ denotes the the deformable attention function, the $F$ denotes the image features and the $\pi$ denotes the transformation from world to pixel coordinates.

    Given the sampled queries $Q_p \in \mathbb{R}^{N \times R \times C}$, we utilize a mixing mechanism following \cite{gao2022adamixer, liu2023sparsebev} to decode and aggregate semantic features and geometric features to obtain a fine-grained Gaussian geometric representation. Specifically, we first predict the dynamic semantic weights $W_s \in \mathbb{R}^{N \times C \times C}$ from Gaussian queries $Q$ based on linear layer, which are used to strengthen the key semantic features of the sampled queries $Q_p$ to get the semantics-aware Gaussian queries $Q_s$, as shown in below:
    \begin{equation}
    Q_s = ReLU(LayerNorm(MatMul(Q_p, W_s)))
    \label{eq6}
    \end{equation}
    Subsequently, we utilize the scale and rotation properties of Gaussian anchors $A$ to predict the dynamic weights of the geometry for describing the fine-grained geometric distribution of Gaussians. Specifically, we concatenate the scale $s$ and rotation $r$ into geometric vectors $G \in \mathbb{R}^{N \times 12}$.  we first predict the dynamic geometric weights $W_g \in \mathbb{R}^{N \times R \times R}$ from geometric vectors $G$ based on linear layer. Then, the geometry-aware Gaussian queries $Q_g$ are generated, as shown in below:
    \begin{equation}
    Q_g = ReLU(LayerNorm(MatMul(W_g, Q_s)))
    \label{eq7}
    \end{equation}

    \textbf{Refinement and Splatting.} We follow the settings of previous works \cite{carion2020end, huang2024gaussianformer, wu2024embodiedocc} to update Gaussian anchors and generate final occupancy predictions. For each property of each Gaussian anchor, the corresponding Gaussian query is utilized to predict the updated vectors with a multi-layer perceptron (MLP). After a few updates, we utilize the Gaussian-to-Voxel splatting to convert sparse Gaussians to dense voxel occupancy for downstream tasks. For loss functions, we train our local occupancy prediction module using the focal loss $L_{focal}$, the lovasz-softmax loss $L_{lov}$, the scene-class affinity loss $L_{geo}$ and $L_{sem}$ following EmbodiedOcc \cite{wu2024embodiedocc}. 
    
\vspace{-2mm}
\section{Experiments}
\label{sec:result}

    In this paper, we propose the RoboOcc model which enhances the geometric and semantic scene understanding for robots. We conduct extensive experiments on Occ-ScanNet and EmbodiedOcc-ScanNet datasets to validate the effectiveness of the proposed method.

\subsection{Datasets}

    \textbf{Occ-ScanNet dataset} \cite{yu2024monocular} consists of 45755 / 19764 frames in the train/val splits. It provides frames  with 12 classes including 1 for free space, and 11 for specific semantics (ceiling, floor, wall, window, chair, bed, sofa, table, tvs, furniture, objects). The annotated voxel grid spans a $4.8m\times4.8m\times2.88m$ box in front of the camera with a resolution of $60\times60\times36$. The local occupancy prediction is trained and evaluated on this dataset. In addition, a mini version of the dataset is available, consists of 5504 / 2376 frames in the train/val splits.
    
    \textbf{EmbodiedOcc-ScanNet dataset} \cite{wu2024embodiedocc} comprises 537 / 137 scenes in the train/val splits. Each scene consists of 30 posed images and corresponding occupancy. The resolutions of global occupancy are calculated by the range of this scene in the world coordinate system with the same voxel size and label categorization as the local prediction task. In global occupancy prediction, we explore the indoor scene sequentially with known camera poses and update the global occupancy prediction via current local observation.

\subsection{Evaluation Metrics}

    Following common practice \cite{cao2022monoscene}, we use mean Intersection-over-Union (mIoU) and Intersection-over-Union (IoU) to evaluate the performance of our model. Specifically, for local occupancy prediction, we perform local evaluation in the front view frustum mask of the current view. For global occupancy prediction, we use the global occupancy of the current scene to compute mIoU and IoU. Global occupancy uses the frustum mask corresponding to 30 frames per scene, which represents the explored regions in the current scene.

\subsection{Implementation Details}

    \textbf{Local Occupancy Prediction.} For image encoder, we use a pre-trained EfficientNet-B7 \cite{tan2019efficientnet} to for multi-scale semantic features. For Gaussian initialization, we use a frozen fine-tuned DepthAnythingV2 model \cite{yang2024depth} to obtain depth-aware features for 3D Gaussians. For hyperparameter settings, the monocular image resolution is set to $480 \times 640$ and the number of Gaussians is 16200 with an upper limit of $0.08m$ on the Gaussians scale. For training settings, We utilize the AdamW \cite{loshchilov2017decoupled} optimizer with a weight decay of 0.01. The learning rate warms up in the first 1000 iterations to a maximum value of 2e-4 and decreases according to a cosine schedule. We train our model for 10 epochs using 8 A100 GPUs on the Occ-ScanNet dataset and 20 epochs on the mini dataset.
    
    \textbf{Global Occupancy Prediction.} We perform further global occupancy prediction based on the pre-training weights obtained from local prediction training. Specifically, we predict the local Gaussian representation at $0.16m$ intervals and update the global occupancy prediction to obtain a global observation of the scene. We train our model for 5 epochs using 8 A100 GPUs on the EmbodiedOcc-ScanNet dataset. The other settings remain the same with the training of the local occupancy prediction.

\begin{table*}[!t]
\renewcommand{\arraystretch}{1.2}
		\caption{
        \textbf{Local Prediction Performance on the Occ-ScanNet dataset.} The state-of-the-art results are marked with \textbf{boldface} and the sub-optimal results are marked with \underline{underline}.
        }
        \small
		\setlength{\tabcolsep}{0.008\textwidth}
        \vspace{-2mm}
        \begin{center}
        \resizebox{1.0\linewidth}{!}{
		\begin{tabular}{l|c|c|c c c c c c c c c c c|c}
			\toprule
			Method
			& Input
			& IoU
			& \rotatebox{90}{\parbox{1.5cm}{\textcolor{ceiling}{$\blacksquare$} ceiling}} 
			& \rotatebox{90}{\textcolor{floor}{$\blacksquare$} floor}
			& \rotatebox{90}{\textcolor{wall}{$\blacksquare$} wall} 
			& \rotatebox{90}{\textcolor{window}{$\blacksquare$} window} 
			& \rotatebox{90}{\textcolor{chair}{$\blacksquare$} chair} 
			& \rotatebox{90}{\textcolor{bed}{$\blacksquare$} bed} 
			& \rotatebox{90}{\textcolor{sofa}{$\blacksquare$} sofa} 
			& \rotatebox{90}{\textcolor{table}{$\blacksquare$} table} 
			& \rotatebox{90}{\textcolor{tvs}{$\blacksquare$} tvs} 
			& \rotatebox{90}{\textcolor{furniture}{$\blacksquare$} furniture} 
			& \rotatebox{90}{\textcolor{objects}{$\blacksquare$} objects} 
			& mIoU\\
            \midrule
            MonoScene~\cite{cao2022monoscene} & $x^{\text{rgb}}$ & 41.60 & 15.17 & 44.71 & 22.41 & 12.55 & 26.11 & 27.03 & 35.91 & 28.32 & 6.57 & 32.16 & 19.84 & 24.62 \\
            ISO~\cite{yu2024monocular} & $x^{\text{rgb}}$ & 42.16 & 19.88 & 41.88 & 22.37 & 16.98 & 29.09 & 42.43 & 42.00 & 29.60 & 10.62 & 36.36 & 24.61 & 28.71 \\
            EmbodiedOcc~\cite{wu2024embodiedocc} & $x^{\text{rgb}}$ & \underline{53.95} & \underline{40.90} & \underline{50.80} & \underline{41.90} & \underline{33.00} & \underline{41.20} & \underline{55.20} & \underline{61.90} & \underline{43.80} & \underline{35.40} & \underline{53.50} & \underline{42.90} & \underline{45.48} \\
            \rowcolor{gray!20}
            RoboOcc~(ours) & $x^{\text{rgb}}$ & \textbf{56.48} & \textbf{45.36} & \textbf{53.49} & \textbf{44.35} & \textbf{34.81} & \textbf{43.38} & \textbf{56.93} & \textbf{63.35} & \textbf{46.35} & \textbf{36.12} & \textbf{55.48} & \textbf{44.78} & \textbf{47.67}\\
            \bottomrule 
            \end{tabular}}
            \end{center}
            \vspace{-2mm}
		  \label{table1}
\end{table*}	

\begin{table*}[!t]
\renewcommand{\arraystretch}{1.2}
		\caption{
        \textbf{Global Prediction Performance on the EmbodiedOcc-ScanNet dataset.} The state-of-the-art results are marked with \textbf{boldface} and the sub-optimal results are marked with \underline{underline}.
        }
        \vspace{-2mm}
		\small
		\setlength{\tabcolsep}{0.008\textwidth}
        \begin{center}
        \resizebox{1.0\linewidth}{!}{
		\begin{tabular}{l|c|c|c c c c c c c c c c c|c}
			\toprule
			Method
			& Input
			& IoU
			& \rotatebox{90}{\parbox{1.5cm}{\textcolor{ceiling}{$\blacksquare$} ceiling}} 
			& \rotatebox{90}{\textcolor{floor}{$\blacksquare$} floor}
			& \rotatebox{90}{\textcolor{wall}{$\blacksquare$} wall} 
			& \rotatebox{90}{\textcolor{window}{$\blacksquare$} window} 
			& \rotatebox{90}{\textcolor{chair}{$\blacksquare$} chair} 
			& \rotatebox{90}{\textcolor{bed}{$\blacksquare$} bed} 
			& \rotatebox{90}{\textcolor{sofa}{$\blacksquare$} sofa} 
			& \rotatebox{90}{\textcolor{table}{$\blacksquare$} table} 
			& \rotatebox{90}{\textcolor{tvs}{$\blacksquare$} tvs} 
			& \rotatebox{90}{\textcolor{furniture}{$\blacksquare$} furniture} 
			& \rotatebox{90}{\textcolor{objects}{$\blacksquare$} objects} 
			& mIoU\\
            \midrule
            SplicingOcc & $x^{\text{rgb}}$ & 49.01 & \textbf{31.60} & 38.80 & 35.50 & 36.30 & 47.10 & 54.50 & 57.20 & 34.40 & 32.50 & 51.20 & 29.10 & 40.74 \\
            EmbodiedOcc~\cite{wu2024embodiedocc} & $x^{\text{rgb}}$ & \underline{51.52} & \underline{22.70} & \textbf{44.60} & \underline{37.40} & \underline{38.00} & \underline{50.10} & \underline{56.70} & \underline{59.70} & \underline{35.40} & \underline{38.40} & \underline{52.00} & \underline{32.90} & \underline{42.53} \\
            \rowcolor{gray!20}
            RoboOcc~(ours) & $x^{\text{rgb}}$ & \textbf{53.30} & 21.94 & \underline{44.57} & \textbf{39.54} & \textbf{38.48} & \textbf{51.28} & \textbf{57.04} & \textbf{63.09} & \textbf{36.70} & \textbf{43.05} & \textbf{54.42} & \textbf{34.38} & \textbf{44.05}\\
            \bottomrule 
            \end{tabular}}
            \end{center}
            \vspace{-2mm}
		  \label{table2}
\end{table*}

\begin{minipage}[!t]{0.4\textwidth}
\renewcommand{\arraystretch}{1.2}
\small
    \centering
    \captionof{table}{
        Ablation on the Components of RoboOcc.
        }
    \label{tab:table3}
    \vspace{-2mm}
    \begin{tabular}{c c|c c}
			\toprule
			GCE
			& OSE
			& IoU
			& mIoU\\
            \midrule
            - & - & 53.93 & 46.20 \\
            \checkmark & - & 54.15 & 46.48 \\
            - & \checkmark & 56.55 & 47.61 \\
            \rowcolor{gray!20}
            \checkmark & \checkmark & \textbf{57.25} & \textbf{47.71}\\
            \bottomrule 
            \end{tabular}
\end{minipage}
\hfill
\begin{minipage}[!t]{0.55\textwidth}
\renewcommand{\arraystretch}{1.2}
\small
    \centering
    \captionof{table}{
        Ablation on the Gaussian Parameters of EmbodiedOcc and RoboOcc.
        }
    \label{tab:table4}
    \vspace{-2mm}
    \resizebox{1.0\linewidth}{!}{
    \begin{tabular}{l |c c|c c}
			\toprule
			Method
			& Number
			& Scale
            & IoU
			& mIoU\\
            \midrule
            \multirow{3}{*}{EmbodiedOcc \cite{wu2024embodiedocc}} & 16200 & 0.08 & 53.93  & 46.20 \\
            & 12150 & 0.16 & 50.26 & 42.65 \\
            & \cellcolor{gray!20} 8100 & \cellcolor{gray!20} 0.20 & \cellcolor{gray!20} 48.81 (-9.49$\%$) & \cellcolor{gray!20} 40.94 (-11.39$\%$) \\
            \midrule
            \multirow{3}{*}{RoboOcc (ours)} & 16200 & 0.08 & 57.25 & 47.71 \\
            & 12150 & 0.16 & 58.73 & 48.92 \\
            & \cellcolor{gray!20} 8100 & \cellcolor{gray!20} 0.20 & \cellcolor{gray!20} 56.21 \textbf{(-1.81$\%$)} & \cellcolor{gray!20} 46.01 \textbf{(-3.56$\%$)}\\
            \bottomrule 
            \end{tabular}}
\end{minipage}

\subsection{Main Results}

\begin{figure*}[!t]
    \centering
        \includegraphics[width=1\textwidth]{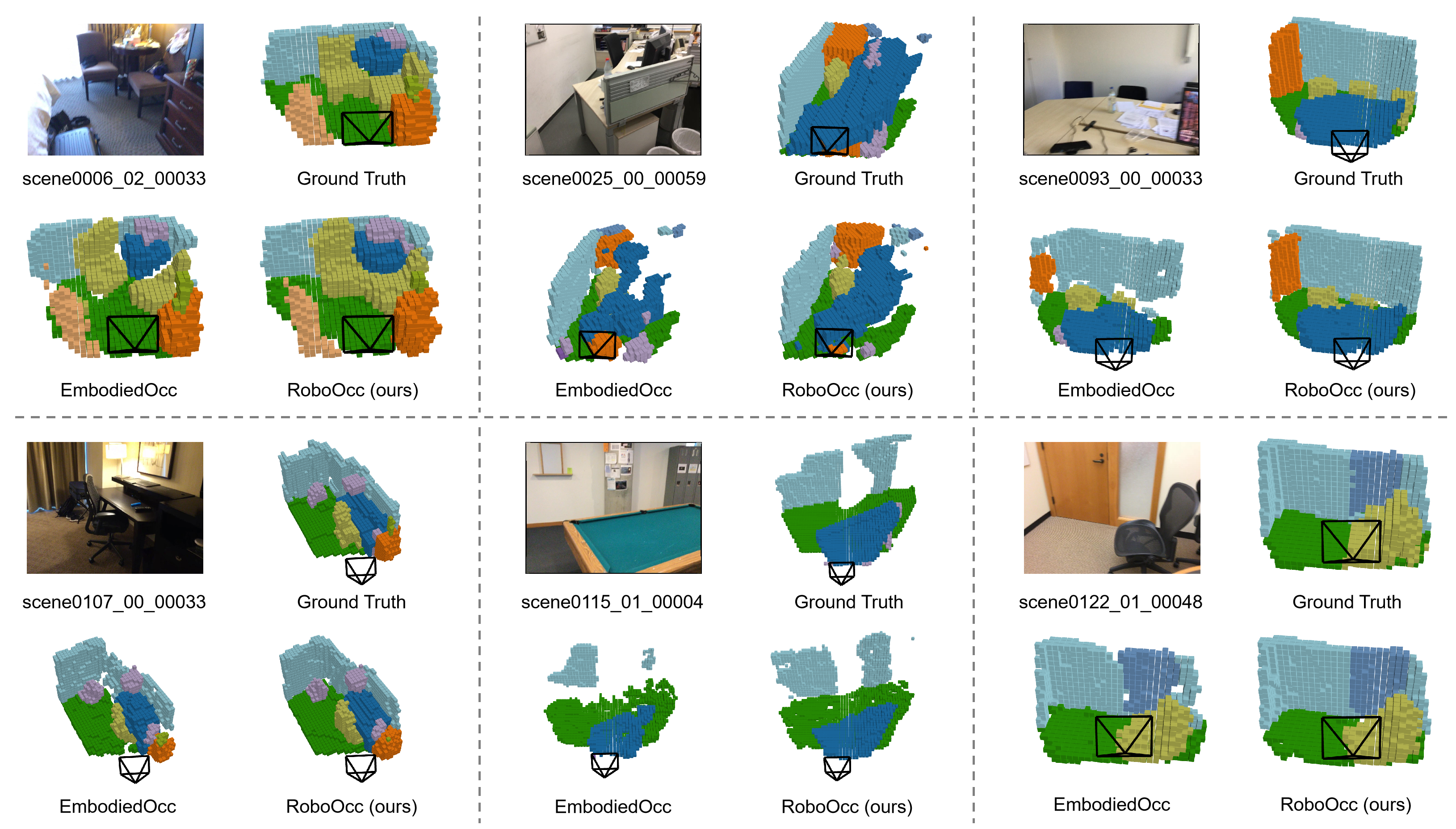}
        \caption{\textbf{Qualitative Analysis on the Occ-ScanNet-mini dataset.} It can be seen that the proposed RoboOcc can model the scene better. It can capture the scene layout and classify various semantic instances more accurately.}
        \label{fig_4}
        \vspace{-2mm}
    \end{figure*}

    \textbf{Local Occupancy Prediction.} We evaluated and compared with existing methods on Occ-ScanNet validation set for local 3D occupancy prediction, as shown in Table \ref{table1}. The proposed RoboOcc outperforms the state-of-the-art method by (2.53, 2.19) in IoU and mIoU metric, respectively. Experimental results demonstrate the strengths of fine-grained modeling and mitigating semantic ambiguity in indoor occupancy prediction. In Figure \ref{fig_4}, we qualitatively analyze the proposed RoboOcc with the existing methods on the Occ-ScanNet-mini dataset.

    \textbf{Global Occupancy Prediction.} We evaluated and compared with existing methods on EmbodiedOcc-ScanNet validation set for global 3D occupancy prediction. The baseline $SplicingOcc$ is obtained by simply splicing the local occupancy prediction sequences. In Table \ref{table2}, the proposed RoboOcc outperforms the state-of-the-art method by (1.78, 1.52) in IoU and mIoU metric, respectively. It can be observed that our RoboOcc demonstrates the generality and extensibility, and also exhibits strength in integrating scene context information.

\subsection{Ablation Study}

    We conduct ablation studies on the Occ-ScanNet-mini dataset to validate the effectiveness of our model design.

    \textbf{Analysis of components of RoboOcc.} In Table \ref{tab:table3}, we provide comprehensive analysis on the components of RoboOcc to validate the effectiveness.We conduct the experiments on Occ-ScanNet-mini validation set, set the number of 3D Gaussians to 16200 and the max scale of 3D Gaussians to $0.08m$. It can be seen that the proposed Geometry-aware Cross-Encoder (GCE) achieved fine-grained scene modeling by accomplishing spatial geometry awareness of 3D Gaussians, which demonstrated the strength on geometric occupancy prediction. On the other hand, the proposed Opacity-guided Self-Encoder (OSE) has a notable influence on the performance, demonstrating that semantic ambiguity is currently a key limitation in indoor occupancy prediction.

    \textbf{Analysis of Gaussian Parameters of RoboOcc.} We analyze the effect of different Gaussian parameters in Table \ref{tab:table4}. It can been seen that decreasing the number and increasing the max scale of the Gaussians can lead to a decrease in performance, because the variations in above parameters cause the increased overlap between Gaussians. 
    The EmbodiedOcc can be seen to decrease the IoU and mIoU by 9.49$\%$ and 11.39$\%$, respectively, at the setting of the maximum parameter variation. On the contrary, the proposed RoboOcc, due to the effective mitigation of semantic ambiguity and the realization of fine-grained modeling in indoor scene, only decrease the IoU and mIoU by 1.81$\%$ and 3.56$\%$, respectively, at the setting of the maximum parameter variation. Further, the proposed RoboOcc surpasses the state-of-the-art method by (8.47, 6.27) in IoU and mIoU metric at the setting of the medium parameter variation, demonstrating the effectiveness of the proposed modules.

\vspace{-2mm}
\section{Conclusion}
\label{sec:conclusion}

    In this paper, considering that current Gaussian representations lack effective utilization of geometric and opacity properties, we propose an enhanced geometric and semantic scene understanding 3D occupancy prediction method for robots, called RoboOcc. Both quantitative and qualitative results have shown that our RoboOcc outperforms existing methods in terms of local occupancy prediction and global occupancy prediction task. We hope our work can shed light on studying more effective scene understanding in indoor occupancy prediction.


\clearpage

\textbf{Limitations.}
Although our model can efficiently estimate the surrounding scene with the enhancement of geometric and semantic scene understanding, semantic learning still faces challenges due to the category imbalance problem (e.g., windows and tvs). In addition, our proposed method only considers 11 common semantic categories in the dataset, which may not fully capture the diversity of categories present in real-world scenes.


\acknowledgments{If a paper is accepted, the final camera-ready version will (and probably should) include acknowledgments. All acknowledgments go at the end of the paper, including thanks to reviewers who gave useful comments, to colleagues who contributed to the ideas, and to funding agencies and corporate sponsors that provided financial support.}


\bibliography{example}  

\newpage
\appendix
\setcounter{page}{1}
\renewcommand{\thepage}{\arabic{page}}
\begin{center}
    {\fontsize{17}{20}\selectfont\textbf{Supplementary Materials}}
\end{center}

\section{Additional Visualizations} 
\begin{figure}[htbp]
    \centering
    \begin{subfigure}[t]{0.48\textwidth}
        \includegraphics[width=\linewidth]{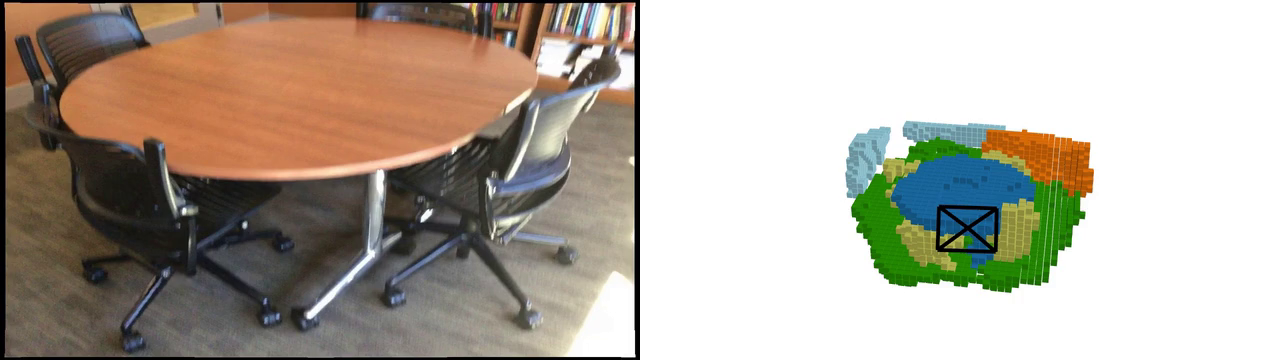}
    \end{subfigure}
    \hfill
    \begin{subfigure}[t]{0.48\textwidth}
        \includegraphics[width=\linewidth]{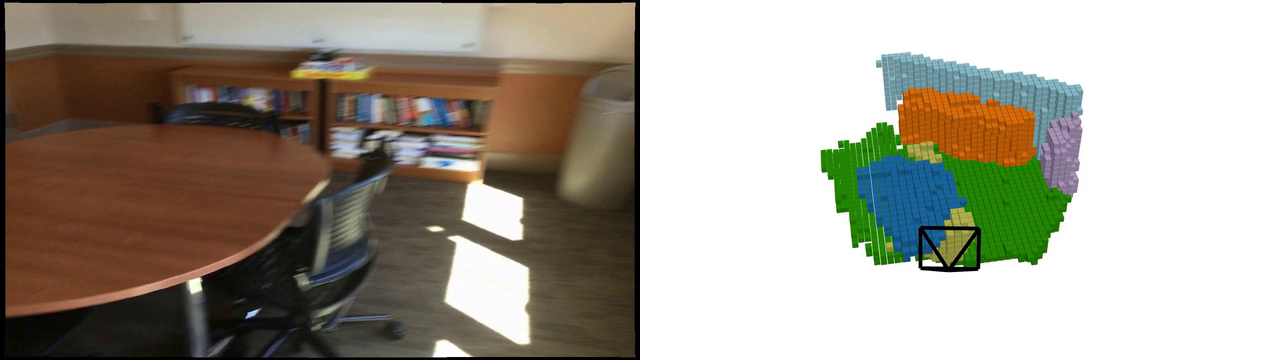}
    \end{subfigure}
    \begin{subfigure}[t]{0.48\textwidth}
        \includegraphics[width=\linewidth]{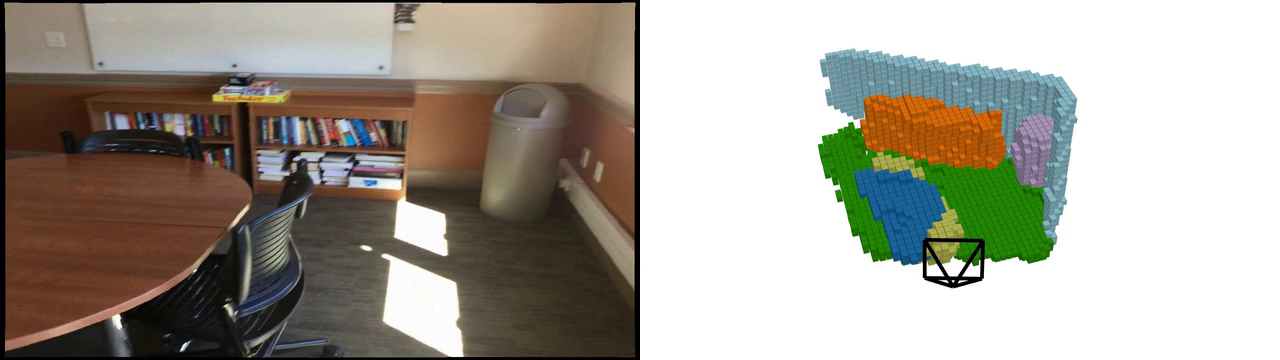}
    \end{subfigure}
    \hfill
    \begin{subfigure}[t]{0.48\textwidth}
        \includegraphics[width=\linewidth]{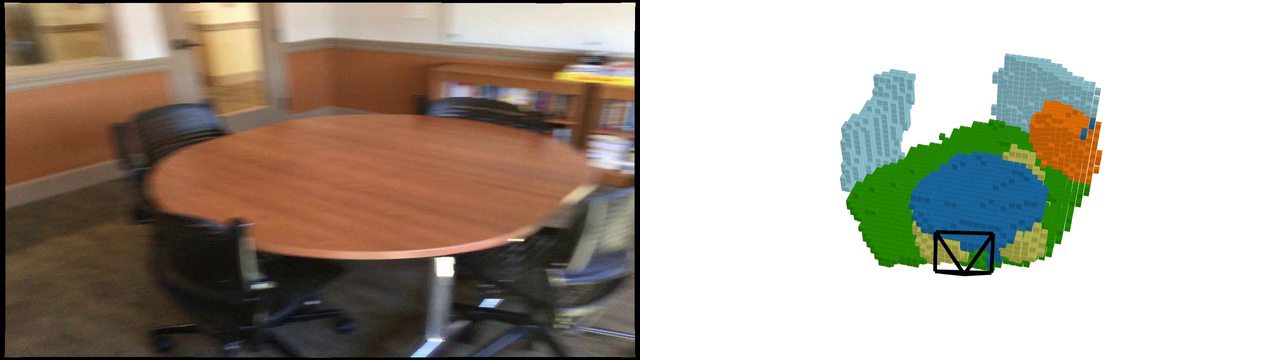}
    \end{subfigure}
    \begin{subfigure}[t]{0.48\textwidth}
        \includegraphics[width=\linewidth]{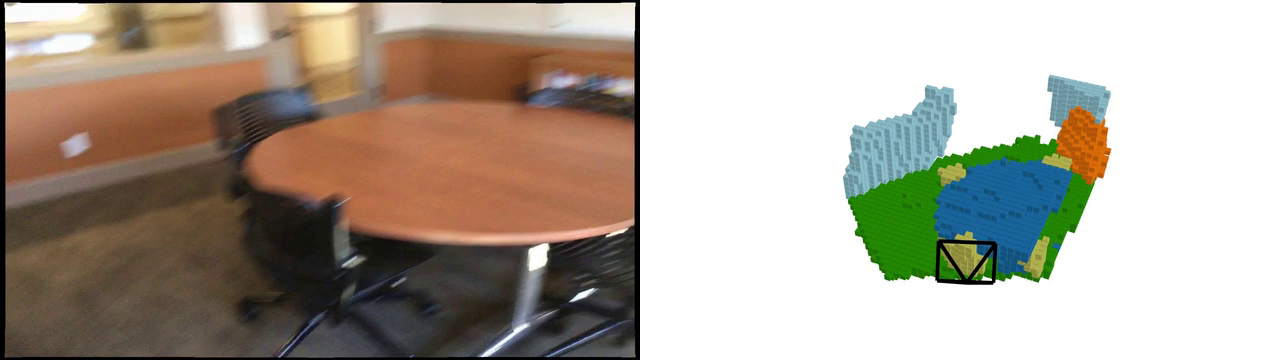}
    \end{subfigure}
    \hfill
    \begin{subfigure}[t]{0.48\textwidth}
        \includegraphics[width=\linewidth]{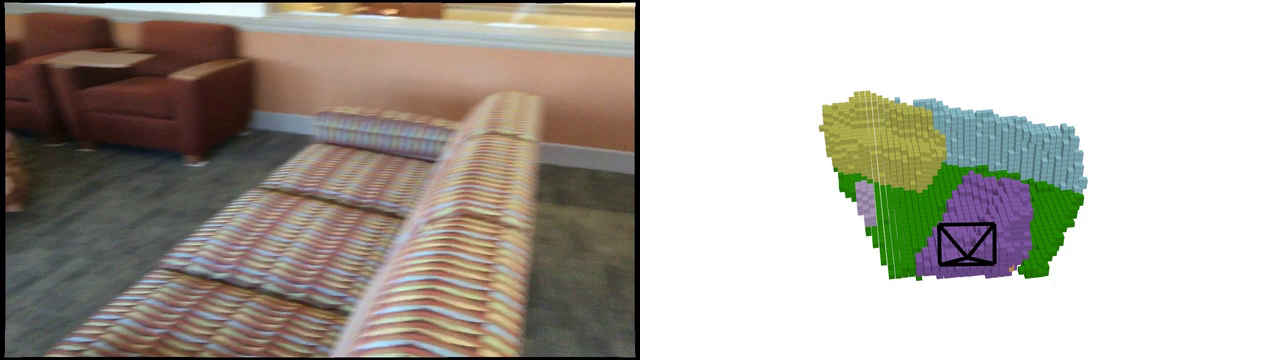}
    \end{subfigure}
    \begin{subfigure}[t]{0.48\textwidth}
        \includegraphics[width=\linewidth]{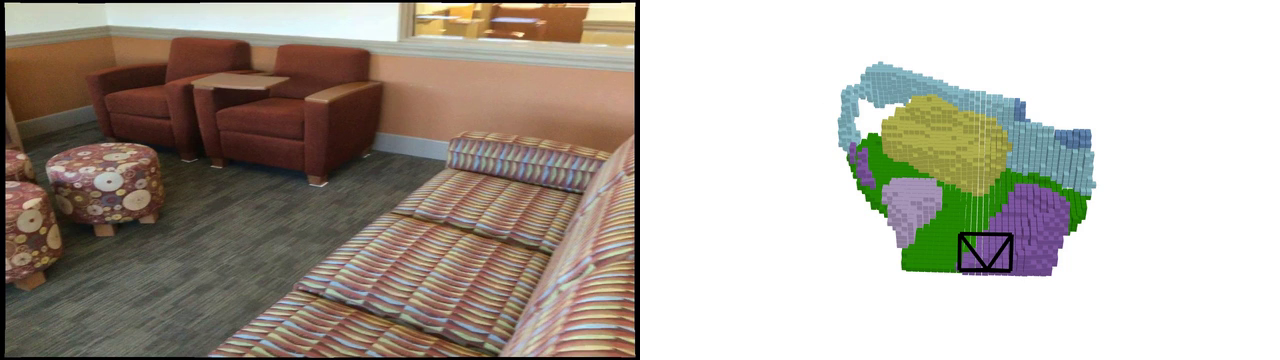}
    \end{subfigure}
    \hfill
    \begin{subfigure}[t]{0.48\textwidth}
        \includegraphics[width=\linewidth]{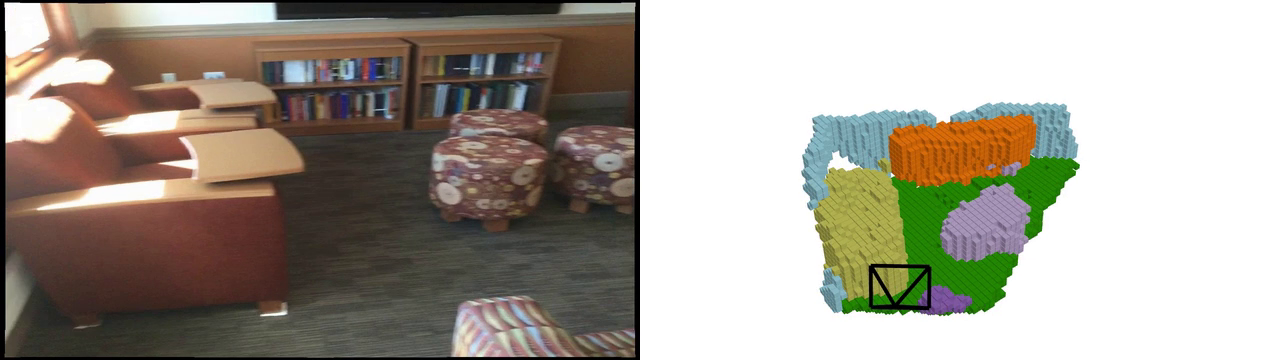}
    \end{subfigure}
    \caption{Additional visualizations in scene0028.}
    \label{fig_5}
\end{figure}

\begin{figure}[htbp]
    \centering
    \begin{subfigure}[t]{0.48\textwidth}
        \includegraphics[width=\linewidth]{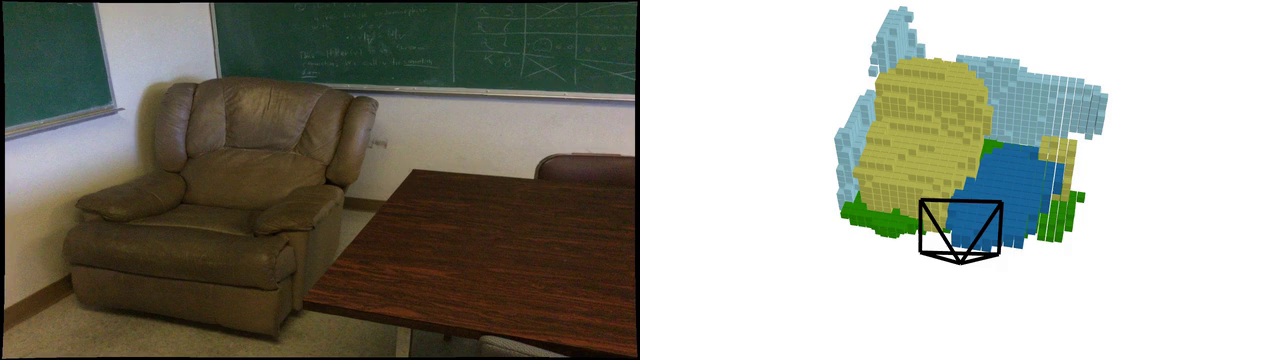}
    \end{subfigure}
    \hfill
    \begin{subfigure}[t]{0.48\textwidth}
        \includegraphics[width=\linewidth]{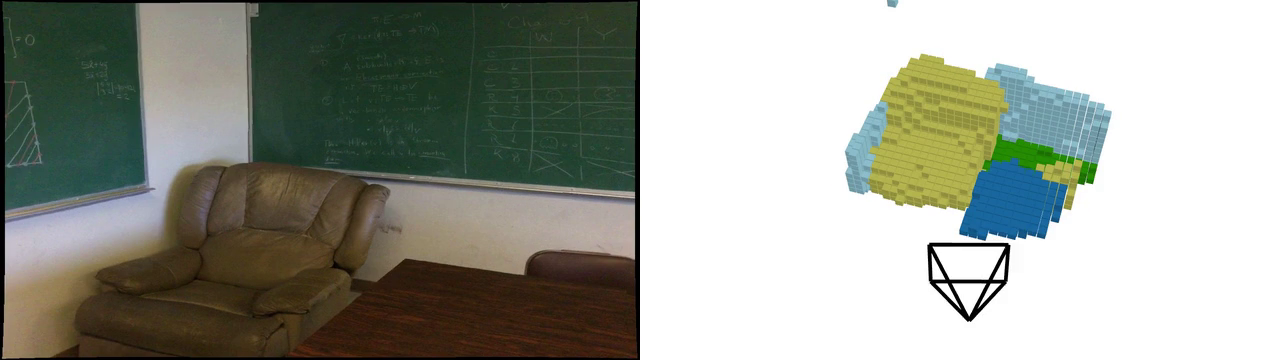}
    \end{subfigure}
    \begin{subfigure}[t]{0.48\textwidth}
        \includegraphics[width=\linewidth]{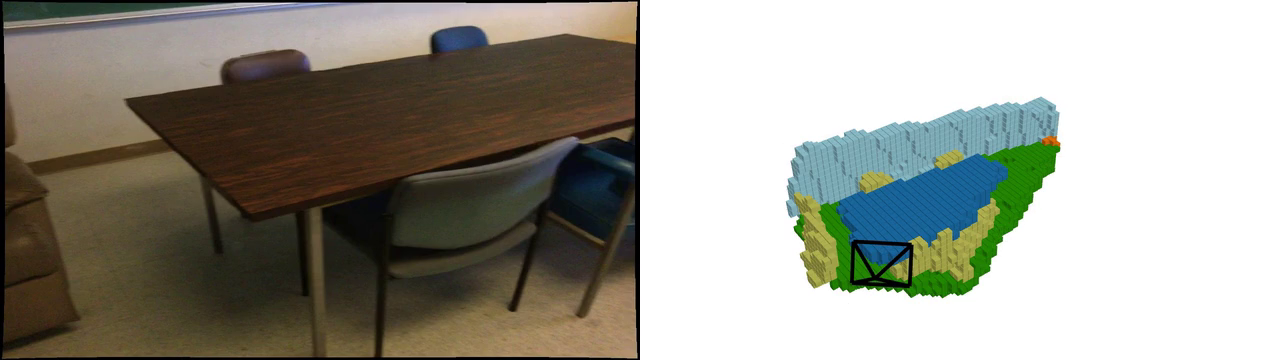}
    \end{subfigure}
    \hfill
    \begin{subfigure}[t]{0.48\textwidth}
        \includegraphics[width=\linewidth]{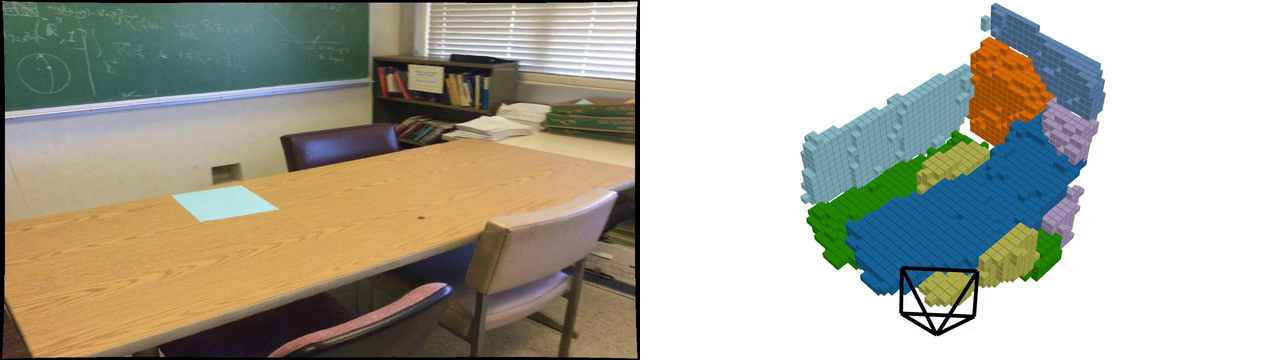}
    \end{subfigure}
    \begin{subfigure}[t]{0.48\textwidth}
        \includegraphics[width=\linewidth]{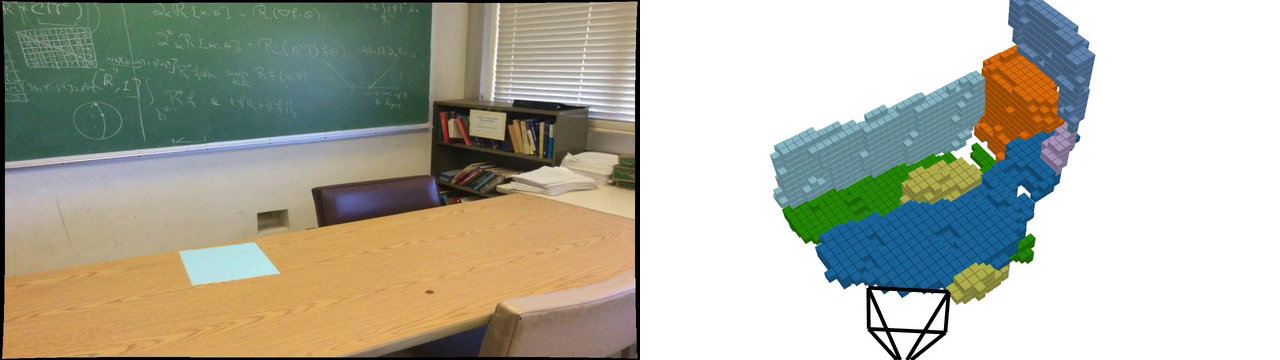}
    \end{subfigure}
    \hfill
    \begin{subfigure}[t]{0.48\textwidth}
        \includegraphics[width=\linewidth]{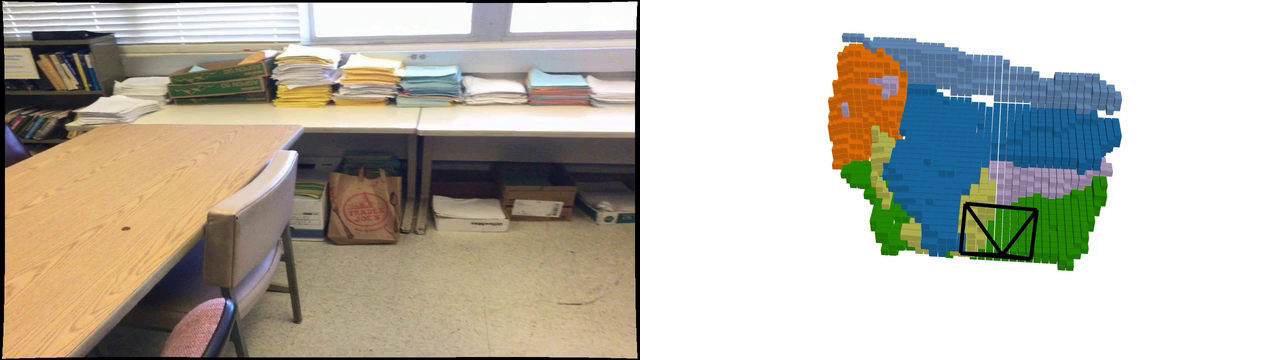}
    \end{subfigure}
    \begin{subfigure}[t]{0.48\textwidth}
        \includegraphics[width=\linewidth]{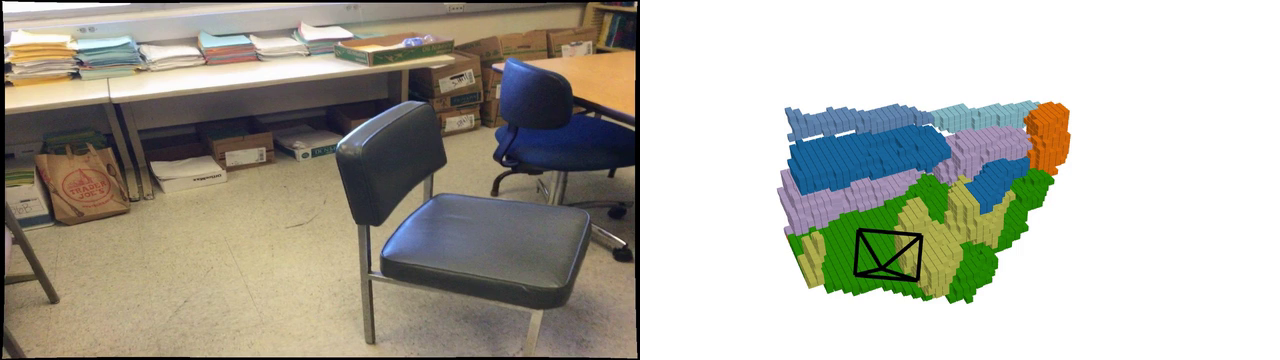}
    \end{subfigure}
    \hfill
    \begin{subfigure}[t]{0.48\textwidth}
        \includegraphics[width=\linewidth]{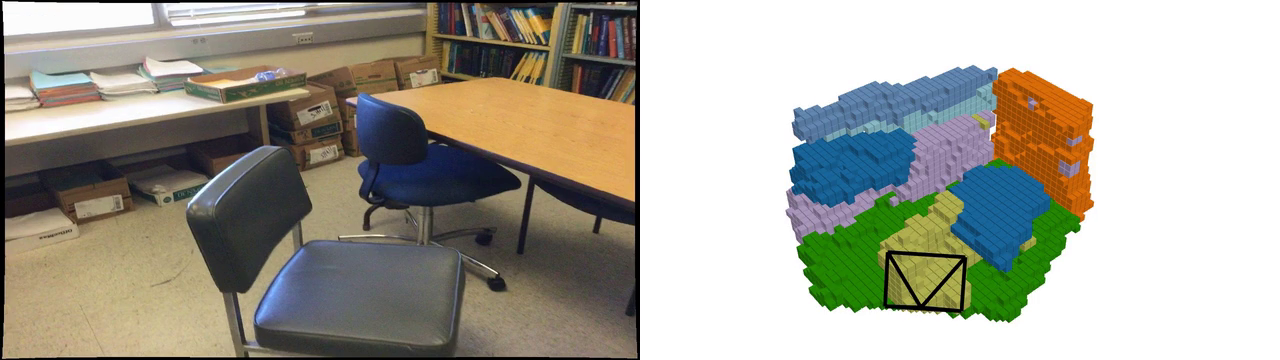}
    \end{subfigure}
    \caption{Additional visualizations in scene0030.}
    \label{fig_6}
\end{figure}

    Figure \ref{fig_5} and \ref{fig_6} show the sampled images from the video demo for 3D occupancy prediction on the Occ-ScanNet \cite{yu2024monocular} validation set. RoboOcc accomplishes fine-grained modeling and performs well in complex scenes as well.

\end{document}